\pdfoutput=1
\documentclass{article}
\PassOptionsToPackage{numbers, compress}{natbib}

\usepackage[main, final]{neurips_2026}
\ifdefined\pdfsuppressptexinfo\pdfsuppressptexinfo=-1\fi

\usepackage[utf8]{inputenc}
\usepackage[T1]{fontenc}
\usepackage{hyperref}
\usepackage{url}
\usepackage{booktabs}
\usepackage{amsfonts}
\usepackage{amssymb}
\usepackage{nicefrac}
\usepackage{microtype}
\usepackage[table]{xcolor}

\usepackage{tabularx}
\usepackage{graphicx}
\usepackage{multirow}
\usepackage{amsmath}
\usepackage{resizegather}
\usepackage{makecell}
\usepackage{placeins}
\usepackage{float}

\title{AdaAlloc: Adaptive Visual Token Allocation for Long-Video Question Answering}

\author{Joungbin An\qquad Kristen Grauman\\
The University of Texas at Austin
}

\begin{document}

\maketitle

\begin{abstract}
Long-video question answering forces multimodal large language models to reason under strict visual-token budgets, creating a tradeoff between broad temporal coverage and fine-grained spatial detail. Uniform sampling preserves coverage but often obscures decisive local cues, while caption-based or frame-similarity retrieval operates over lossy proxies that can miss motion, state changes, and before--after relations. We recast long-video reasoning as a \emph{visual-token allocation} problem: given a fixed budget, decide how much to spend on global context versus high-resolution local evidence, and where in the video to draw that local evidence from. We introduce AdaAlloc, a training-free inference framework that adaptively allocates visual tokens between a global video overview and localized visual detail. AdaAlloc first plans a global/local allocation policy, then uses temporal grounding to locate candidate evidence segments and refines them into a compact temporal memory. The selected policy controls both frame sampling density and resolution, combining sparse, lower-resolution global frames with denser, higher-resolution local frames drawn from the refined memory. The final answerer reasons over the allocated global context and refined local evidence. Across LVBench, MLVU, VideoMME, and LongVideoBench, AdaAlloc improves over budget-matched uniform sampling by up to 10.5 percentage points and outperforms the compared frame-selection and agentic methods on their reported benchmarks.\footnote{Project webpage: \url{https://vision.cs.utexas.edu/projects/adaalloc}.}

\end{abstract}

\section{Introduction}

Long videos pose a challenge for multimodal large language models (MLLMs) operating under a fixed visual-token budget, as highlighted by recent long-video question answering benchmarks and MLLM studies~\citep{mangalam2023egoschema,wang2025lvbench,zhou2025mlvu,fu2025video,wu2024longvideobench}. Answering a question often requires tracking a narrative over many minutes, yet the decisive clue may be a number on a screen, a brief hand motion, or a subtle state change. A sparse sweep of the whole video preserves temporal coverage but blurs such details. A dense look at a few moments preserves detail, but helps only if those moments are the ones needed to correctly answer the question. Effective long-video understanding therefore requires deciding not just which frames to show, but \emph{how many tokens to spend where}.

Existing work treats this primarily as a selection or compression problem. Uniform sampling preserves coverage at the cost of resolution, while query-aware keyframe methods improve relevance but remain constrained by a frame budget~\citep{tang2025adaptive,yu2024self,huang2025frag,liu2025bolt,zhu2025focus,chen2026eventanchored}. Retrieval and agentic methods locate relevant clips, captions, or memories~\citep{zhang2023simple,wang2024videoagent,wang2025videotree,arnab2025temporal,luo2025videorag,shen2025vgent,zuo2025videolucy}, yet many rely on caption-based indexing or frame-wise similarity, which can underrepresent motion and fine-grained visual cues. Recent advances in adaptive compression, query-type routing, multimodal keyframe selection, and temporal grounding improve which evidence reaches the model~\citep{wang2025adaretake,zhang2025flexselect,li2025divide,wang2026wheretofocus,zhang2025vtimecot,shen2025zoomzero}, but they leave open how to distribute representation capacity across both time and resolution.

We recast long-video reasoning as a \emph{visual-token allocation} problem:
given a fixed budget,
how should tokens be divided between low-resolution global coverage and high-resolution local inspection?
This is distinct from frame selection: the objective is to allocate fidelity, not just to choose frames.
Intuitively, some questions are global-dominant, such as summarization, event ordering, or identifying the overall activity; these often benefit more from broad temporal coverage than from high-resolution detail at every frame, and can therefore be represented by lower-resolution, lower-token-per-frame views.
Other questions are local-dominant, such as reading text, counting objects, or distinguishing a brief hand action, where the answer depends on fine-grained visual evidence and would benefit from denser tokens at those time points.
Many questions require both: global context to situate the episode and local evidence to resolve the answer.

This problem formulation yields two coupled decisions. First, how should the budget be split between global and local? Second, conditioned on a local budget, where should those tokens be placed? The second decision is critical. Spending local tokens on top-ranked frames can fail when retrieval is noisy, when similar events repeat, or when the query requires a before--after comparison rather than a single salient moment.
Local allocation must therefore identify the right temporal evidence---not just salient frames, but candidate intervals that contain the events, context, and transitions needed to answer the question.

We introduce AdaAlloc, a training-free framework for budget-aware long-video reasoning. Starting from a sparse visual overview of the full video, AdaAlloc runs a structured evidence-planning step to identify the question's evidence needs, select one of five global/local allocation policies---global-only, global-heavy, balanced, local-heavy, or local-only---and determine whether targeted temporal search is needed. When local evidence is required, it generates evidence-seeking queries and invokes a lightweight video temporal grounding model (we use UniversalVTG~\citep{an2026universalvtg}) to propose candidate evidence regions. These proposals, together with timestamps inferred from the question or overview, form a temporal memory. Because grounding outputs can be noisy, redundant, or fragmented, AdaAlloc further refines this memory with the MLLM, filtering and consolidating candidates conditioned on the question and evidence plan. The final visual input follows the selected policy: a global stream of sparse frames over the full video, and a local stream of denser, higher-resolution frames drawn from the refined memory. In this way, AdaAlloc decides both how broadly to look and where to look closely.

\begin{figure}[t!]
  \centering
  \includegraphics[width=0.93\textwidth]{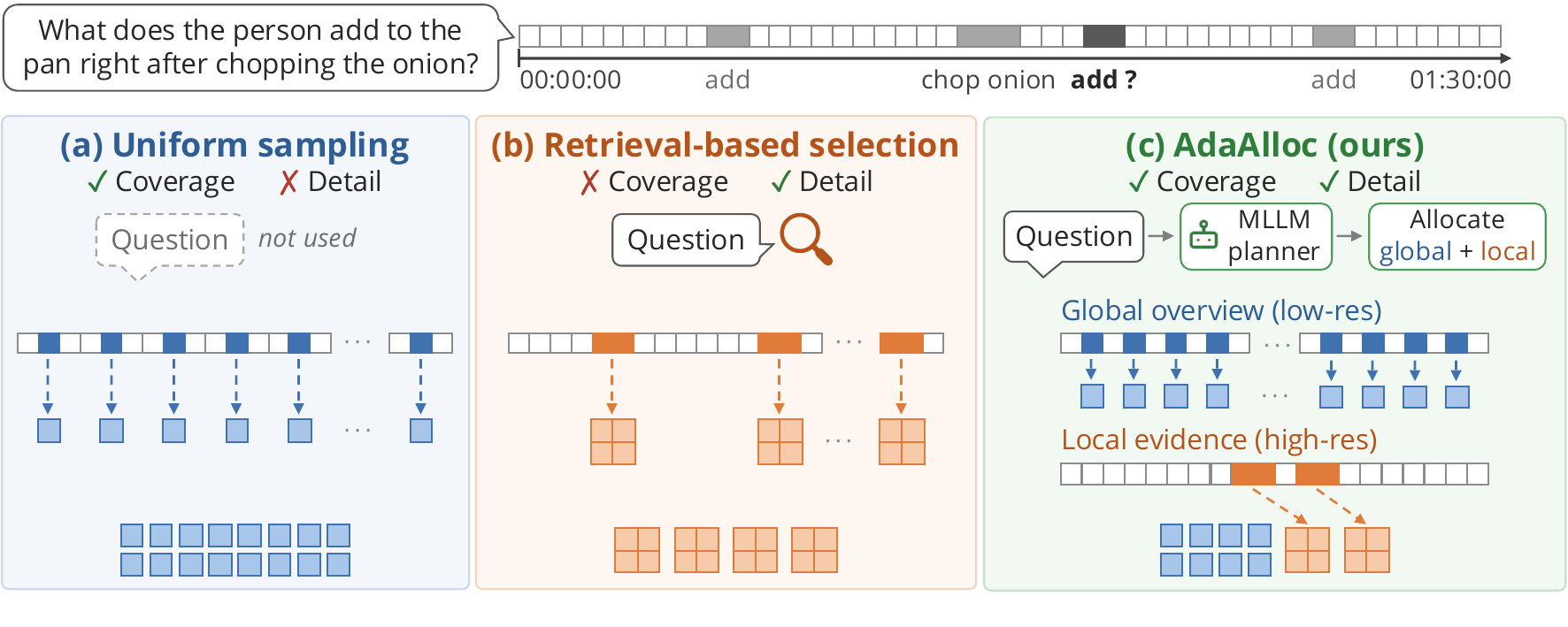}
\vspace{-3mm}
\caption{
\textbf{Visual token allocation for long-video reasoning.}
Uniform sampling preserves global coverage but loses fine details, while retrieval-based selection focuses on a few clips but can lose context. AdaAlloc allocates a fixed token budget between low-resolution global coverage and high-resolution local evidence for both context and fine-grained verification.
}
  \label{fig:teaser}
  \vspace{-2mm}
\end{figure}

This view subsumes two common extremes: uniform sampling~\citep{maaz2023videochatgpt,zhang2024longva} is a global-only policy that preserves coverage but underspecifies detail, while retrieval-only~\citep{huang2025frag,sun2025f2c} is a local-only policy that concentrates detail but risks losing context. See Figure~\ref{fig:teaser}.
AdaAlloc makes this tradeoff question-conditioned. Experiments on LVBench~\cite{wang2025lvbench}, MLVU~\cite{zhou2025mlvu}, VideoMME~\cite{fu2025video}, and LongVideoBench~\cite{wu2024longvideobench} show that, under the same token budget and MLLM backbone, AdaAlloc improves over uniform sampling by up to 10.5 percentage points, with the largest gains on the longest-video benchmarks. Improvements on both Qwen3-VL-8B and VideoLLaMA3-7B support its use as a plug-and-play, training-free agentic framework for improving existing MLLMs' long-video understanding.

Our contributions are:
\begin{itemize}
    \item We formulate long-video understanding as visual-token allocation under a fixed budget, making explicit the tradeoff between temporal coverage and spatial detail.
    \item We introduce AdaAlloc, a training-free agentic framework that jointly determines how to split a fixed visual-token budget between global coverage and local detail, and where to spend the local share, using overview-conditioned planning, temporal grounding, and MLLM-based memory refinement.
\end{itemize}

\section{Related Work}
\label{sec:related_work}
\paragraph{Long-video MLLMs under token constraints.}
Recent long-video benchmarks~\citep{mangalam2023egoschema,wang2025lvbench,zhou2025mlvu,fu2025video,wu2024longvideobench} expose a shared constraint: evidence is dispersed over minutes or hours, but MLLMs can process only a limited number of visual tokens at adequate resolution. Existing methods typically address this constraint from one of two directions. One line preserves broad temporal coverage by extending context windows or compressing visual tokens~\citep{zhang2024longva,li2024llama,shen2024longvu,song2024moviechat,wang2025adaretake,zhang2025flexselect}. This helps maintain a global view of the video, but can lose fine-grained evidence when many frames must share a fixed token budget. Another line selects a subset of frames or clips to improve relevance under a strict budget~\citep{yu2024self,tang2025adaptive,huang2025frag,liu2025bolt,zhu2025focus,yu2024frame,sun2025mdp3,guo2025logic,ye2025re,chen2026eventanchored, ramakrishnan2023spotem}. These methods can preserve local detail, but their success depends on the selected evidence being both correct and sufficient, and they may discard the global context needed for ordering, repeated events, or disambiguation. Query-adaptive variants further show that the sampling strategy should depend on the question~\citep{li2025divide,wang2026wheretofocus}, and related ideas have been explored for allocating text reasoning tokens in LLMs~\citep{li2025selfbudgeter}. AdaAlloc builds on this insight but changes the decision: rather than choosing between global coverage and local detail, it allocates a fixed visual-token budget between them, deciding when to see broadly, when to inspect locally, and when both are needed.

\paragraph{Evidence seeking for long-video reasoning.}
A second line of work treats long-video reasoning as evidence seeking: first identify relevant video regions, then answer from the selected evidence. Caption-based systems convert frames or clips into textual memories for retrieval~\citep{zhang2023simple,kahatapitiya2024language,ma2024drvideo}, while agentic and retrieval-augmented methods use planning, tools, captions, entities, or graph-structured memories to decide what evidence to inspect~\citep{min2024morevqa,wang2024videoagent,fan2024videoagent,wang2025videotree,arnab2025temporal,luo2025videorag,ren2025videorag,jeong2025videorag,shen2025vgent,zuo2025videolucy,wang2025active}.
SpotEM~\citep{ramakrishnan2023spotem} learns to select promising clips using low-cost semantic features, concentrating expensive feature extraction on query-relevant regions.
Recent agents learn this search behavior with supervised fine-tuning and/or reinforcement learning: VSeek~\citep{goel2026vseek} learns to issue natural-language search queries and reason over the retrieved clips using verifiable rewards, while LOVE-R1~\citep{fu2025lover1} and Video-o3~\citep{zeng2026videoo3} start from a low-resolution view of the whole video and, over multiple turns, inspect selected clips at higher resolution. AdaAlloc is an entirely training-free framework that jointly allocates a fixed visual-token budget between global temporal coverage and local high-resolution evidence. Unlike LOVE-R1 and Video-o3, which add high-resolution clips to a fixed global view, AdaAlloc spends less on global context whenever it spends more on local detail. Overview-conditioned planning determines this budget split, while temporal grounding and MLLM-based memory refinement determine where the local share is spent, producing a single budget-matched visual input for final answering.

\paragraph{Temporal grounding as a mechanism, not a task.}
Video temporal grounding (VTG) localizes language queries in untrimmed video~\citep{gao2017tall,lei2021detecting,lin2023univtg,snag,an2025hieramamba,an2026universalvtg}. Recent time-aware MLLMs aim to improve the model's own timestamp awareness, through temporal instruction tuning, visual timelines, reinforcement learning, or zoom-in verification~\citep{ren2023timechat,huang2023vtimellm,qian2024momentor,nie2024slowfocus,wang2024grounded,zhang2025vtimecot,zhang2026timelens,wang2025time,tvg-r1,li2025videochat,feng2025video,shen2025zoomzero,li2025universal}.
Instead, AdaAlloc assigns temporal localization a different role: rather than making timestamp prediction the reasoning objective, it uses VTG as an inference-time tool for visual-token allocation. It decides when localized evidence is needed, grounds evidence-seeking queries with UniversalVTG~\citep{an2026universalvtg}, and refines the resulting intervals into a temporal memory before spending high-resolution tokens. Unlike methods that primarily decide \emph{which} frames or segments to process~\citep{li2025divide,shen2025zoomzero}, AdaAlloc also decides how much global context to preserve and where local detail is worth the budget.

\section{Problem Formulation}
\label{sec:problem_formulation}

We study long-video multiple-choice question answering under a fixed visual-token budget. Let $V$ be an untrimmed video of duration $T$, $q$ a natural-language question with answer choices $\mathcal{A}=\{a_1,\ldots,a_m\}$, and $B$ the maximum number of visual tokens available to the MLLM. The goal is to predict $\hat{a}\in\mathcal{A}$ from a visual representation $E(V,q;B)$ whose total visual-token count does not exceed $B$. This count depends on both the number of sampled frames and their encoded spatial resolution. Encoding the same frames at higher resolution requires more tokens but can preserve finer visual details. A fixed token budget therefore couples temporal coverage and spatial detail, requiring decisions about both how many frames to sample and at what resolution to encode them.

Throughout, $B$ denotes the visual-token budget of a single MLLM call. Every MLLM call in AdaAlloc, including planning and memory refinement (Section~\ref{sec:AdaAlloc}), is capped at the same $B$, and our controlled comparisons match $B$ for the final answering call that receives $E(V,q;B)$. In our experiments, we instantiate $B$ with 16K and 32K visual-token regimes.\footnote{Including planning and refinement, AdaAlloc's end-to-end runtime is $2.3$--$2.5\times$ that of single-pass uniform sampling, with an average accuracy gain of $4.8$--$5.3$ percentage points across the four benchmarks. Appendix~\ref{sec:appendix_cost} reports total compute and runtime for all inference stages.}%

We decompose the evidence into a global overview stream $E_g$ and a local detail stream $E_l$. The global stream samples the full video to preserve temporal context, such as scene changes, event order, repeated episodes, and coarse narrative structure. The local stream allocates denser or higher-resolution frames to selected temporal regions, preserving fine-grained cues such as text, counts, attributes, spatial relations, brief actions, and state changes. The framework chooses both a budget split $\pi=(\rho_g,\rho_l)$
\[
\rho_g,\rho_l \in [0,1], \qquad \rho_g+\rho_l=1, \qquad B_g=\rho_gB,\quad B_l=\rho_lB,
\]
and, when $B_l>0$, a set of local intervals $\mathcal{R}=\{(s_j,e_j)\}_{j=1}^K$. The final MLLM input is
\[
E(V,q;B)=\mathrm{concat}\!\left(E_g(V;B_g), E_l(V,\mathcal{R};B_l)\right).
\]

This formulation separates two decisions that are often conflated: \emph{how much} budget should remain global versus local, and \emph{where} local budget should be placed. Uniform sampling is the global-only endpoint, preserving coverage but limiting detail. Retrieval-only pipelines approximate the local-only endpoint, concentrating budget on selected intervals but risking loss of context needed for temporal order, repeated events, or disambiguation.

\section{AdaAlloc Approach}
\label{sec:AdaAlloc}

\begin{figure}[t!]
  \centering
  \includegraphics[width=\linewidth]{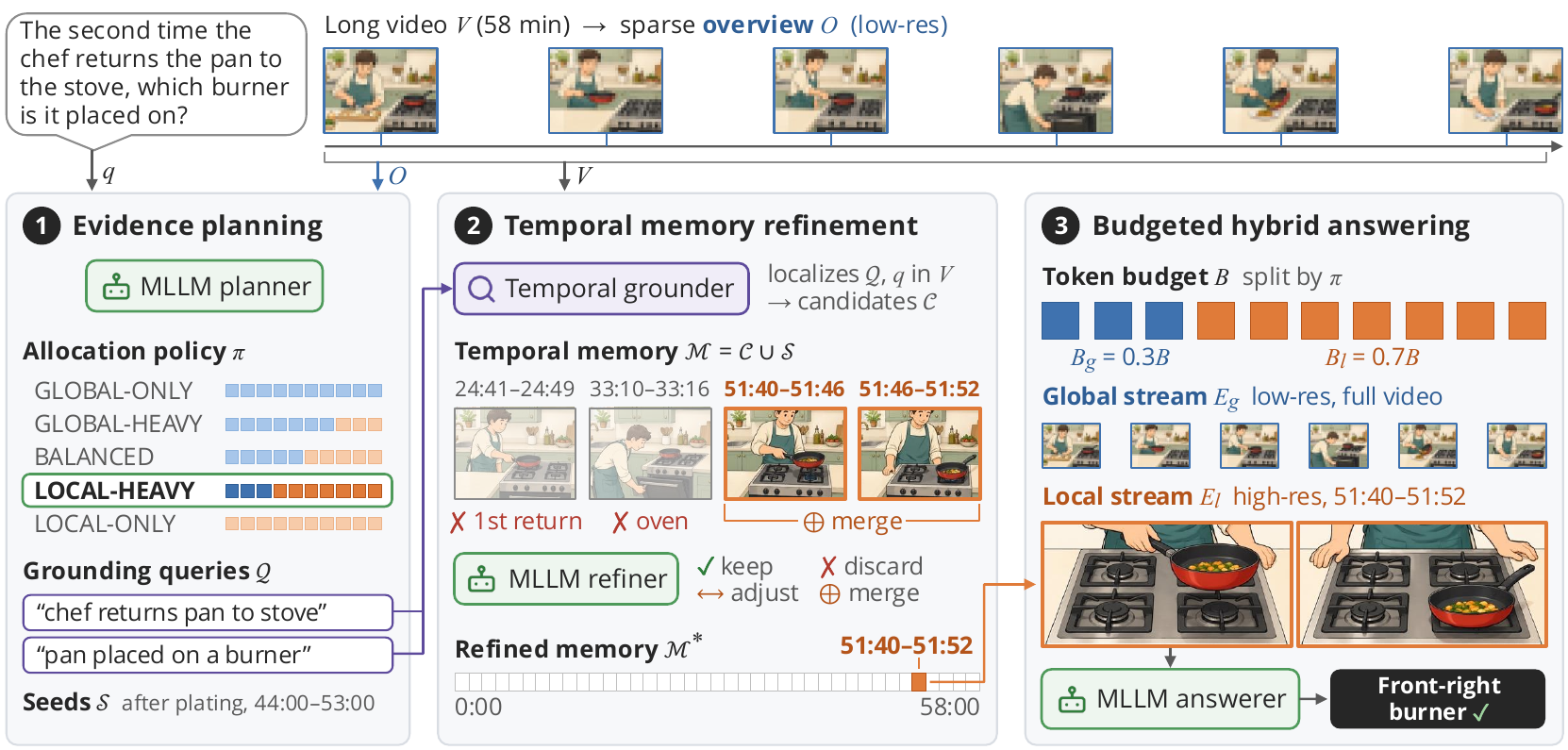}
  \vspace{-4mm}
\caption{
\textbf{AdaAlloc pipeline.}
AdaAlloc splits a fixed visual-token budget $B$ between low-res global context (blue) and high-res local evidence (orange).
\textbf{(1)}~From a sparse overview $O$, an MLLM planner picks a policy $\pi$ and generates queries $\mathcal{Q}$.
\textbf{(2)}~A temporal grounder (UniversalVTG) localizes $\mathcal{Q}$; the MLLM refines the candidates into $\mathcal{M}^*$, discarding an off-topic clip and the \emph{first} return to the stove.
\textbf{(3)}~The MLLM answers from a global stream $E_g$ and a local stream $E_l$ drawn from $\mathcal{M}^*$.
}
  \label{fig:main_results_visual}
  \vspace{-2mm}
\end{figure}

Building on the formulation in Section~\ref{sec:problem_formulation}, AdaAlloc instantiates the evidence representation
$E(V,q;B)$ with an adaptive three-stage pipeline, shown in Figure~\ref{fig:main_results_visual}.
\textbf{First}, overview-conditioned planning samples a sparse low-resolution overview of $V$ and uses it to select a global/local allocation policy $\pi=(\rho_g,\rho_l)$ and generate evidence-seeking temporal grounding queries. These queries are used to retrieve candidate intervals from the original video. \textbf{Second}, temporal-memory refinement consolidates the retrieved intervals, explicit timestamps, and overview-derived temporal seeds into a compact refined memory $\mathcal{M}^*$. \textbf{Third}, budgeted hybrid answering constructs the final representation
\[
E(V,q;B)=\mathrm{concat}\!\left(E_g(V;B_g), E_l(V,\mathcal{R}^*;B_l)\right),
\]
where $B_g=\rho_gB$, $B_l=\rho_lB$, and $\mathcal{R}^*$ denotes the local intervals selected from $\mathcal{M}^*$.
The model then answers from this hybrid input, using $E_g$ for global temporal context and $E_l$ for high-resolution inspection of the selected evidence regions.

\subsection{Overview-conditioned evidence planning}
\label{sec:overview_planning}

AdaAlloc begins by sampling a sparse, low-resolution overview $O$ of the full video $V$. It captures coarse information about the setting, actors, scene progression, repeated events, and approximate temporal anchors to guide evidence acquisition and global/local allocation. The final answerer then receives a question-specific combination of global context and local detail within $B$ (Section~\ref{sec:hybrid_answering}). Planning and refinement respect the same per-call budget $B$.
Conditioned on the overview, question, and answer choices, the MLLM planner produces
\[
(\pi, \mathcal{Q}, \mathcal{S}) = f_{\mathrm{plan}}(O, q, \mathcal{A}),
\]
where $\pi$ is the allocation policy, $\mathcal{Q}$ is a set of grounding queries, and $\mathcal{S}$ is a set of temporal memory seeds.

The allocation policy $\pi=(\rho_g,\rho_l)$ is selected from five discrete options:
\[
\pi \in
\left\{
\begin{array}{ll}
\textsc{Global-only}  & (1.0, 0.0) \\
\textsc{Global-heavy} & (0.7, 0.3) \\
\textsc{Balanced}    & (0.5, 0.5) \\
\textsc{Local-heavy} & (0.3, 0.7) \\
\textsc{Local-only}  & (0.0, 1.0)
\end{array}
\right\},
\]
where each pair denotes the fraction of the visual-token budget assigned to global overview frames and local evidence clips, respectively. These five discrete policies simplify the planner's decision while allowing the allocation to adapt to each video--question pair. In practice, selecting among them matches or exceeds the average accuracy of a variant of our approach that directly predicts the global/local ratio (Appendix~\ref{sec:appendix_robustness}).

The query set $\mathcal{Q}$ is generated when the planner determines that temporal search is needed. Each query describes visually grounded evidence relevant to $q$, such as an action, object interaction, state change, repeated event, or temporal anchor. In parallel, the planner may produce temporal seeds $\mathcal{S}$ directly from the question or overview, including explicit timestamps, intervals around overview-visible events, or coarse windows likely to contain relevant before/after context. The resulting $(\pi,\mathcal{Q},\mathcal{S})$ defines the allocation intent and temporal search space passed to the memory refinement stage.
Our framework builds in some tolerance to imperfect planning and seeds: grounding and memory refinement can correct inaccurate temporal seeds, while our overview-density analysis shows limited aggregate sensitivity to sparse planning input (Appendix~\ref{sec:appendix_robustness}).

\subsection{Temporal memory construction and refinement}
\label{sec:memory_refinement}

Given the planner outputs $(\pi,\mathcal{Q},\mathcal{S})$, AdaAlloc constructs a temporal memory when the policy assigns nonzero budget to local evidence. For each grounding query in $\mathcal{Q}$, together with the original question $q$ as a complementary query, AdaAlloc invokes UniversalVTG~\citep{an2026universalvtg} over the original video. UniversalVTG is a lightweight $\sim$60M-parameter VTG model that localizes open-vocabulary queries directly in video time, producing ranked candidate intervals before costly MLLM inspection. We choose UniversalVTG for its strong grounding performance and efficiency, though alternative VTG models could be dropped in, and AdaAlloc with simple query--frame similarity retrieval also outperforms uniform sampling (Appendix~\ref{sec:appendix_robustness}). We denote the resulting VTG candidates as $\mathcal{C}$.

AdaAlloc initializes the temporal memory by combining VTG candidates with planner-derived seeds:
\[
\mathcal{M} = \mathcal{C} \cup \mathcal{S}.
\]
Each entry $m_i \in \mathcal{M}$ stores a temporal span $[s_i,e_i]$, source provenance, the associated query or rationale, and the VTG retrieval score for grounded candidates or the planner's confidence for temporal seeds.
The memory is constructed for recall rather than precision: VTG predictions, overview-derived seeds, and explicit timestamps may be shifted, redundant, fragmented, or irrelevant. AdaAlloc therefore treats $\mathcal{M}$ as a set of timestamped hypotheses, not as final evidence.

Before spending the local budget $B_l$, AdaAlloc refines the initialized memory with:
\[
\mathcal{M}^* = f_{\mathrm{refine}}(\mathcal{M}, O, q, \mathcal{A}).
\]
For each candidate interval, the model observes sampled frames together with the question, answer choices, planner rationale, source provenance, associated query, and VTG score when available. It then edits the memory using four operations: \emph{keep} useful evidence regions, \emph{discard} off-topic or duplicate candidates, \emph{adjust} shifted intervals, and \emph{merge} adjacent or complementary regions that capture one event, repeated action, or before/after context.

The output $\mathcal{M}^*$ is a compact set of timestamped regions with provenance and sampling priority. This stage uses temporal grounding for candidate generation, not final evidence selection: UniversalVTG provides high-recall temporal hypotheses, which the model refines according to the planned evidence needs. The refined memory determines the local intervals $\mathcal{R}^*$ used in the final budgeted representation.

\subsection{Budgeted hybrid answering}
\label{sec:hybrid_answering}

AdaAlloc assembles the final visual input from the selected allocation policy and refined memory $\mathcal{M}^*$. The global budget is spent on sparse, low-resolution frames spanning the full video, while the local budget is spent on denser, higher-resolution clips sampled from $\mathcal{M}^*$. When multiple memory regions survive refinement, the local budget is divided across the selected regions; if no reliable local memory remains, it falls back to additional overview coverage.

This produces a hybrid representation with a global stream for temporal context and detail streams for localized inspection. Global questions rely primarily on the overview, localized questions allocate more budget to grounding-guided evidence clips, and mixed-evidence questions preserve context while reserving local capacity for decisive cues. The final answerer reasons over this hybrid input, turning the refined memory into an explicit allocation of visual tokens across time and resolution. The full pipeline improves accuracy over uniform sampling across all four benchmarks under matched final-answering budgets, with its end-to-end accuracy--cost tradeoff quantified in Section~\ref{sec:problem_formulation} and Appendix~\ref{sec:appendix_cost}.

\begin{table}[t!]
\centering
\scriptsize
\setlength{\tabcolsep}{3pt}
\caption{Summary of the long-form video QA benchmarks used in our evaluation.
}
\vspace{-2mm}
\label{tab:benchmarks}
\begin{tabular*}{\columnwidth}{@{\extracolsep{\fill}}lrrrrrl}
\toprule
\textbf{Benchmark} & \textbf{\#Video} & \textbf{\#Question} & \textbf{Avg. (min.)} & \textbf{Range (min.)} & \textbf{Total Hours} & \textbf{Main Stressor} \\
\midrule
LVBench~\citep{wang2025lvbench}
& 103 & 1,549 & 67.3 & 30.1--139.9 & 115.5 & Hour-scale context \\
MLVU-Test~\citep{zhou2025mlvu}
& 349 & 502 & 17.8 & 3.1--136.8 & 103.4 & Hard six-way QA \\
VideoMME~\citep{fu2025video}
& 900 & 2,700 & 17.0 & 0.18--59.7 & 255.3 & Duration robustness \\
LongVideoBench-Val~\citep{wu2024longvideobench}
& 753 & 1,337 & 7.9 & 0.13--59.6 & 99.8 & Long-context referring \\
\bottomrule
\end{tabular*}
\vspace{-3mm}
\end{table}

\section{Experiments}

\paragraph{Benchmarks.}
We evaluate AdaAlloc on four complementary long-form video question-answering benchmarks, summarized in Table~\ref{tab:benchmarks}.
\textbf{LVBench}~\citep{wang2025lvbench} targets extreme long-video comprehension, with hour-scale videos that stress global narrative understanding and
sparse evidence retrieval. \textbf{MLVU-Test}~\citep{zhou2025mlvu} is the newer and harder official test split of MLVU: compared with MLVU-Dev, it
expands multiple-choice questions from four to six options and adds new categories such as Sports QA and Tutorial QA, making chance performance lower and
multi-task reasoning more challenging. \textbf{VideoMME}~\citep{fu2025video} provides a balanced evaluation across short, medium, and long videos,
testing whether a method remains robust across duration regimes rather than only on extreme long videos. \textbf{LongVideoBench}~\citep{wu2024longvideobench}
focuses on long-context video-language reasoning with referring questions that require precise localization of relevant
visual evidence; we report results on the validation split. Together, these benchmarks test dense local evidence retrieval, global temporal
comprehension, temporal ordering, counting, OCR/detail perception, and long-context referring reasoning across diverse video lengths and domains. All
benchmarks are evaluated in a video-only setting without subtitles, so models must answer from visual evidence rather than transcript cues.

\begin{table}[t]
\centering
\small
\setlength{\tabcolsep}{4pt}
\renewcommand{\arraystretch}{1.08}
\caption{
\textbf{Controlled main results under matched visual-token budgets.}
Following the protocol in Section~\ref{sec:implementation_details}, all methods use the same backbone, decoding setup, and visual-token budget within each block.
We report accuracy (\%) on four long-form video QA benchmarks in the video-only setting.
Numbers in parentheses denote absolute gains over uniform sampling under the same budget.
}
\label{tab:main_results}
\begin{tabular*}{\textwidth}{@{\extracolsep{\fill}}llcccc@{}}
\toprule
\textbf{Budget} & \textbf{Method}
& \textbf{LVBench}
& \textbf{MLVU-Test}
& \textbf{VideoMME$_{\text{w/o sub}}$}
& \textbf{LongVideoBench-Val} \\
\midrule
\multirow{2}{*}{16K tokens}
& Uniform sampling
& 42.7 & 55.4 & 67.1 & 62.8 \\
& \textbf{AdaAlloc}
& \textbf{51.7} \textcolor{teal}{\scriptsize (+9.0)}
& \textbf{61.2} \textcolor{teal}{\scriptsize (+5.8)}
& \textbf{68.4} \textcolor{teal}{\scriptsize (+1.3)}
& \textbf{66.0} \textcolor{teal}{\scriptsize (+3.2)} \\
\midrule
\multirow{2}{*}{32K tokens}
& Uniform sampling
& 46.1 & 61.4 & 69.5 & 63.9 \\
& \textbf{AdaAlloc}
& \textbf{56.6} \textcolor{teal}{\scriptsize (+10.5)}
& \textbf{67.1} \textcolor{teal}{\scriptsize (+5.7)}
& \textbf{70.7} \textcolor{teal}{\scriptsize (+1.2)}
& \textbf{67.5} \textcolor{teal}{\scriptsize (+3.6)} \\
\bottomrule
\end{tabular*}
\end{table}

\vspace{-2mm}
\paragraph{Implementation details.}
\label{sec:implementation_details}
We implement AdaAlloc with Qwen3-VL-8B-Instruct~\citep{bai2025qwen3} and use the same model as the final answerer for the main controlled baselines.
Following the fixed-budget formulation in Section~\ref{sec:problem_formulation}, our controlled comparisons use matched final-answering visual-token budgets.
We report two budget regimes, 16K and 32K visual tokens, corresponding to uniform-sampling reference settings of 64 and 128 frames, respectively.
These settings cover a practical range for current long-video MLLM inference with 7--8B-class models: the 64-frame setting represents a constrained regime where allocation is especially important, while the 128-frame setting gives uniform sampling substantially stronger temporal coverage while remaining feasible on typical 40GB-class GPUs in our setup.

The uniform baseline spends the entire budget on sparse global coverage: it samples the video at 1 FPS and uniformly subsamples up to the corresponding frame limit.
Retrieval-based baselines spend the same budget on localized clips returned by temporal retrieval.
AdaAlloc receives the same total budget, but distributes it between a global overview stream and local detail streams according to one of the five policies defined in Section~\ref{sec:overview_planning}: \textsc{Global-only}, \textsc{Global-heavy}, \textsc{Balanced}, \textsc{Local-heavy}, and \textsc{Local-only}.
When uniform full-video sampling yields sufficiently dense temporal coverage under the assigned budget, AdaAlloc answers directly. These routed examples are answered from uniformly sampled global frames with the standard multiple-choice QA prompt, bypassing VTG queries, temporal proposals, memory construction, and refinement.
For VTG-guided candidate generation, we use UniversalVTG~\citep{an2026universalvtg} and retain the top-5 temporal proposals per query before temporal-memory refinement.
Frames in the overview and local streams are dynamically sampled and resized to satisfy the selected global/local visual-token split, using lower-resolution frames for broad overview coverage and higher-resolution frames for local inspection.
Within each stream, the allocated budget and sampled frame count determine the available visual tokens per frame and hence the spatial resolution. Frames are resized accordingly, preserving aspect ratio with height and width aligned to multiples of 32 pixels. All MLLM calls use greedy decoding.

\paragraph{Main results.}
Table~\ref{tab:main_results} compares methods under matched backbone, decoding setup, and visual-token budgets. Using Qwen3-VL-8B-Instruct as a fixed backbone isolates the effect of visual-token allocation. Uniform or fixed-rate frame sampling is a common inference strategy for VideoLLMs and a standard baseline for adaptive frame selection and efficient long-video understanding methods~\citep{maaz2023videochatgpt,lin2023videollava,zhang2024longva,tang2025adaptive,yu2024frame,zhang2025qframe,liu2025bolt}. AdaAlloc consistently improves over this standard sampling baseline across all four benchmarks and both budget regimes, with the largest gains on LVBench~\cite{wang2025lvbench} and MLVU-Test~\cite{zhou2025mlvu}. These benchmarks contain the longest videos in our evaluation, where answer-relevant evidence is often sparse and temporally localized.
Under a fixed token budget, uniform sampling may miss temporally localized evidence or capture it at insufficient spatial resolution to resolve fine details.
AdaAlloc addresses these limitations by retaining a global overview while using temporal grounding to allocate high-resolution inspection to likely evidence regions. The gains are smaller but consistent on VideoMME~\cite{fu2025video} (duration and sub-task breakdowns in Appendix~\ref{sec:appendix_videomme}) and LongVideoBench-Val~\cite{wu2024longvideobench}, suggesting that adaptive allocation remains useful even when uniform sampling already provides stronger coverage. Overall, these results support our central claim: under a fixed visual-token budget, long-video QA benefits from allocating visual fidelity across time rather than spending the entire budget uniformly.

\paragraph{Comparison with same-backbone selection methods.}
Table~\ref{tab:frame_selection_comparison} compares AdaAlloc with recent frame-selection methods under the same Qwen3-VL-8B-Instruct backbone.
Because methods differ in visual preprocessing, equal frame counts do not imply equal visual-token budgets, so we compare at approximate token budgets. DIG~\citep{li2025divide} reports results at fixed frame counts, with each frame encoded into approximately 150 visual tokens. Exact 16K and 32K results are not reported for prior methods, so we use the nearest reported settings, 128 and 256 frames ($\approx$19K and $\approx$38K tokens), which give AKS and DIG more visual tokens than AdaAlloc.
Despite the smaller budget, AdaAlloc outperforms AKS~\citep{tang2025adaptive} and DIG~\citep{li2025divide} at both operating points, with the largest gains on the VideoMME medium subset (+10.3 and +9.0 points over DIG).
These results suggest that AdaAlloc's benefit is not merely from selecting a better subset of frames, but from allocating a fixed visual budget between global temporal context and high-resolution local inspection.

\begin{table}[t!]
\centering
\small
\setlength{\tabcolsep}{5pt}
\renewcommand{\arraystretch}{1.08}
\caption{
\textbf{Comparison with recent frame-selection methods under the same backbone.}
All methods use Qwen3-VL-8B-Instruct. $^\dagger$Approximate budgets converted from the reported 128- and 256-frame settings at about 150 visual tokens per frame~\citep{li2025divide}; these exceed AdaAlloc's 16K and 32K budgets, making the budget favorable to the baselines. We report LongVideoBench and the medium/long subsets of VideoMME$_{\text{w/o sub}}$ to match DIG~\citep{li2025divide}.
}
\label{tab:frame_selection_comparison}
\begin{tabular}{lcccc}
\toprule
\multirow{2}{*}{\textbf{Method}}
& \multirow{2}{*}{\begin{tabular}[c]{@{}c@{}}\textbf{Visual}\\\textbf{Tokens}\end{tabular}}
& \multirow{2}{*}{\begin{tabular}[c]{@{}c@{}}\textbf{LongVideo}\\\textbf{Bench-Val}\end{tabular}}
& \multicolumn{2}{c}{\textbf{VideoMME$_{\text{w/o sub}}$}} \\
\cmidrule(lr){4-5}
& & & \textbf{Medium} & \textbf{Long} \\
\midrule
AKS~\citep{tang2025adaptive}
& $\approx$19K$^\dagger$ & 53.0 & 54.4 & 51.1 \\
DIG~\citep{li2025divide}
& $\approx$19K$^\dagger$ & 58.3 & 58.4 & 51.1 \\
\textbf{AdaAlloc}
& 16K & \textbf{66.0} & \textbf{68.7} & \textbf{57.2} \\
\midrule
AKS~\citep{tang2025adaptive}
& $\approx$38K$^\dagger$ & 55.6 & 59.2 & 53.2 \\
DIG~\citep{li2025divide}
& $\approx$38K$^\dagger$ & 61.2 & 61.6 & 53.8 \\
\textbf{AdaAlloc}
& 32K & \textbf{67.5} & \textbf{70.6} & \textbf{62.2} \\
\bottomrule
\end{tabular}
\vspace{-3mm}
\end{table}

\begin{table*}[t!]
\centering
\small
\setlength{\tabcolsep}{4.5pt}
\renewcommand{\arraystretch}{1.08}
\caption{
\textbf{Contextual comparison with representative long-video understanding systems.}
Prior numbers are taken from the respective papers when available.
We include methods that report results on comparable long-video benchmarks and specify a frame budget or sampling setting for similar-scale systems.
Visual inputs are reported in their original units, since frame-to-token conversion depends on the backbone, resolution, and compression scheme.
Because methods differ in backbone MLLM, training data, inference pipeline, visual preprocessing, and evaluation protocol, this table is intended as a cross-paper comparison rather than a strictly controlled head-to-head evaluation.
Bold marks the best result excluding proprietary models.
}
\label{tab:external_comparison}
\resizebox{\textwidth}{!}{
\begin{tabular}{lcccccc}
\toprule
\textbf{Method} & \textbf{Size} & \textbf{Visual input}
& \textbf{LVBench}
& \textbf{MLVU-Test}
& \textbf{VideoMME$_{\text{w/o sub}}$}
& \textbf{LongVideoBench-Val} \\
\midrule
\multicolumn{7}{l}{\textit{Proprietary models}} \\
GPT-4o~\citep{hurst2024gpt4o}
& -- & 384 frames & 48.9 & 54.9 & 71.9 & 66.7 \\
Gemini-1.5-Pro~\citep{team2024gemini15}
& -- & 256 frames & 33.1 & 53.8 & 75.0 & 64.0 \\

\midrule
\multicolumn{7}{l}{\textit{Open-source long-video MLLMs / trained or adapted video models}} \\
mPLUG-Owl3~\citep{ye2024mplugowl3}
& 7B & 128 frames & 43.5 & -- & 59.3 & 52.1 \\
NVILA~\citep{liu2024nvila}
& 8B & 256 frames & -- & -- & 64.2 & 57.7 \\
ZoomV~\citep{pan2025zoomv}
& 8B & 80 frames & 51.6 & -- & 65.6 & 63.6 \\
TimeMarker~\citep{chen2024timemarker}
& 8B & 128 frames & -- & -- & 57.3 & 56.3 \\
LongVILA~\citep{chen2024longvila}$^\S$
& 7B & 256 frames & -- & 49.0 & 60.1 & 57.1 \\
SF-LLaVA-1.5~\citep{sfllava2025}
& 7B & 128 frames & -- & -- & 63.9 & 62.5 \\
LongVA~\citep{zhang2024longva}
& 7B & 128 frames & -- & -- & 52.6 & -- \\
MemVid~\citep{yuan2025memory}
& 7B & 128 frames & 44.4 & 58.1 & 63.7 & -- \\

\midrule
\multicolumn{7}{l}{\textit{Training-free / inference-time selection and retrieval methods}} \\
BOLT~\citep{liu2025bolt}$^\S$
& 8B & 32 frames & -- & -- & 65.7 & 62.2 \\
F2C~\citep{sun2025f2c}
& 7B & 32 frames & -- & -- & 65.6 & 60.8 \\
VideoDetective~\citep{yang2026videodetective}
& 8B & 32 frames & 43.2 & 56.3 & -- & 60.2 \\
QuoTA~\citep{luo2025quota}
& 7B & 64 frames & -- & -- & 65.9 & 59.0 \\
Video-RAG~\citep{luo2025videorag}
& 7B & 64 frames & -- & -- & -- & 58.7 \\
Q-Frame~\citep{zhang2025qframe}$^\S$
& 7B & 64 frames & -- & -- & 64.5 & 59.9 \\
FOCUS~\citep{zhu2025focus}
& 7B & 64 frames & -- & -- & 65.4 & 63.5 \\
\midrule
\rowcolor{gray!10}
AdaAlloc
& 8B & 16K tokens$^\ddagger$ & 51.7 & 61.2 & 68.4 & 66.0 \\
\rowcolor{gray!10}
AdaAlloc
& 8B & 32K tokens$^\ddagger$ & \textbf{56.6} & \textbf{67.1} & \textbf{70.7} & \textbf{67.5} \\
\bottomrule
\end{tabular}
}
\par\vspace{2pt}
{\footnotesize\raggedright
$^\ddagger$At approximately 256 visual tokens per frame, 16K and 32K tokens correspond to \textbf{64} and \textbf{128} frames, respectively. AdaAlloc adapts frame counts and resolutions within these budgets. Other methods are listed using their reported frame counts.
$^\S$From other sources: BOLT (Qwen3-VL-8B) from its official GitHub; Q-Frame (LLaVA-Video-7B) from FOCUS~\citep{zhu2025focus}; LongVILA MLVU-Test from MemVid~\citep{yuan2025memory}.\par}
\vspace{-1mm}
\end{table*}

\paragraph{Comparison with existing systems.}
Table~\ref{tab:external_comparison} situates AdaAlloc among representative long-video understanding systems.
We focus on methods that report results on comparable benchmarks and provide a frame budget or sampling setting for similar-scale models.
Because prior methods differ in backbone, training data, inference pipeline, visual preprocessing, and evaluation protocol, this table should be read as a cross-paper comparison rather than as a strictly controlled head-to-head comparison.
AdaAlloc achieves competitive accuracy without additional model training at 16K and 32K visual tokens, roughly equivalent to 64 and 128 frames at 256 tokens per frame.
Compared with training-free and inference-time selection methods, AdaAlloc performs favorably while using a unified allocation mechanism that preserves global overview context and selectively spends high-resolution tokens on temporally grounded local evidence.

\begin{table}[t!]
\centering
\small
\setlength{\tabcolsep}{4pt}
\renewcommand{\arraystretch}{1.08}
\caption{
\textbf{Ablation study under the 32K visual-token budget.}
All variants use Qwen3-VL-8B-Instruct with the same final-answering visual-token budget.
Uniform sampling answers from uniformly sampled frames.
Retrieval-only answers from segments localized by UniversalVTG using the original question.
Fixed policy uses a \textsc{Balanced} global/local split.
W/o temporal memory refinement removes the refinement stage (Section~\ref{sec:memory_refinement}).
}
\vspace{-1mm}
\label{tab:ablation}
\begin{tabular*}{\textwidth}{@{\extracolsep{\fill}}lcccc@{}}
\toprule
\textbf{Method}
& \textbf{LVBench}
& \textbf{MLVU-Test}
& \textbf{VideoMME$_{\text{w/o sub}}$}
& \textbf{LongVideoBench-Val} \\
\midrule
Uniform sampling
& 46.1 & 61.4 & 69.5 & 63.9 \\
Retrieval-only
& 48.2 & 59.4 & 67.6 & 63.5 \\
Fixed policy
& 55.0 & 60.6 & 69.0 & 64.7 \\
w/o temporal memory refinement
& 55.9 & 66.5 & 70.2 & 67.2 \\
\midrule
\textbf{AdaAlloc}
& \textbf{56.6} & \textbf{67.1} & \textbf{70.7} & \textbf{67.5} \\
\bottomrule
\end{tabular*}
\vspace{-4mm}
\end{table}

\paragraph{Ablations.}
Table~\ref{tab:ablation} isolates the main design choices in AdaAlloc under the 32K visual-token budget. Uniform sampling follows a common inference setup for base video MLLMs, in which the model answers directly from frames sampled uniformly across the video. Retrieval-only grounds the original question with UniversalVTG~\cite{an2026universalvtg} and answers from frames sampled from the resulting temporal proposals. Fixed policy keeps the same global/local input structure as AdaAlloc but replaces question-conditioned allocation with a fixed \textsc{Balanced} split. W/o temporal memory refinement removes the refinement stage from Section~\ref{sec:memory_refinement} and samples directly from all the candidate proposals. AdaAlloc performs best across all benchmarks, indicating that its gains come from combining global context, adaptive allocation, and refined local evidence selection.
A complementary 16K ablation in Appendix~\ref{sec:appendix_ablation16k} separates planning, the global/local hybrid input, and adaptive allocation: adding a fixed global stream to a local-only variant changes accuracy only slightly, whereas making the split question-conditioned improves all four benchmarks by 1.5--3.2 points.

\paragraph{Generalization, robustness, and cost.}
AdaAlloc's gains are not tied to a specific backbone or grounding model (Appendix~\ref{sec:appendix_robustness}). Applied to VideoLLaMA3-7B~\citep{zhang2025videollama3} under the same 32K budget, it improves over uniform sampling on all four benchmarks (+2.25 points on average), and replacing UniversalVTG with query--frame similarity retrieval retains most of the gains (0.31 average points higher at 16K and 1.66 lower at 32K).
Its selected policies also follow interpretable question-type patterns, with global-oriented policies more common for descriptive questions (what or where) than temporal questions (before or after) (Appendix~\ref{sec:appendix_policy}).
The additional planning and refinement calls cost more than single-pass uniform sampling, but AdaAlloc still has lower latency than DIG~\citep{li2025divide} at both budgets (Appendix~\ref{sec:appendix_cost}).

\begin{figure}[t!]
  \centering
  \includegraphics[width=1\textwidth]{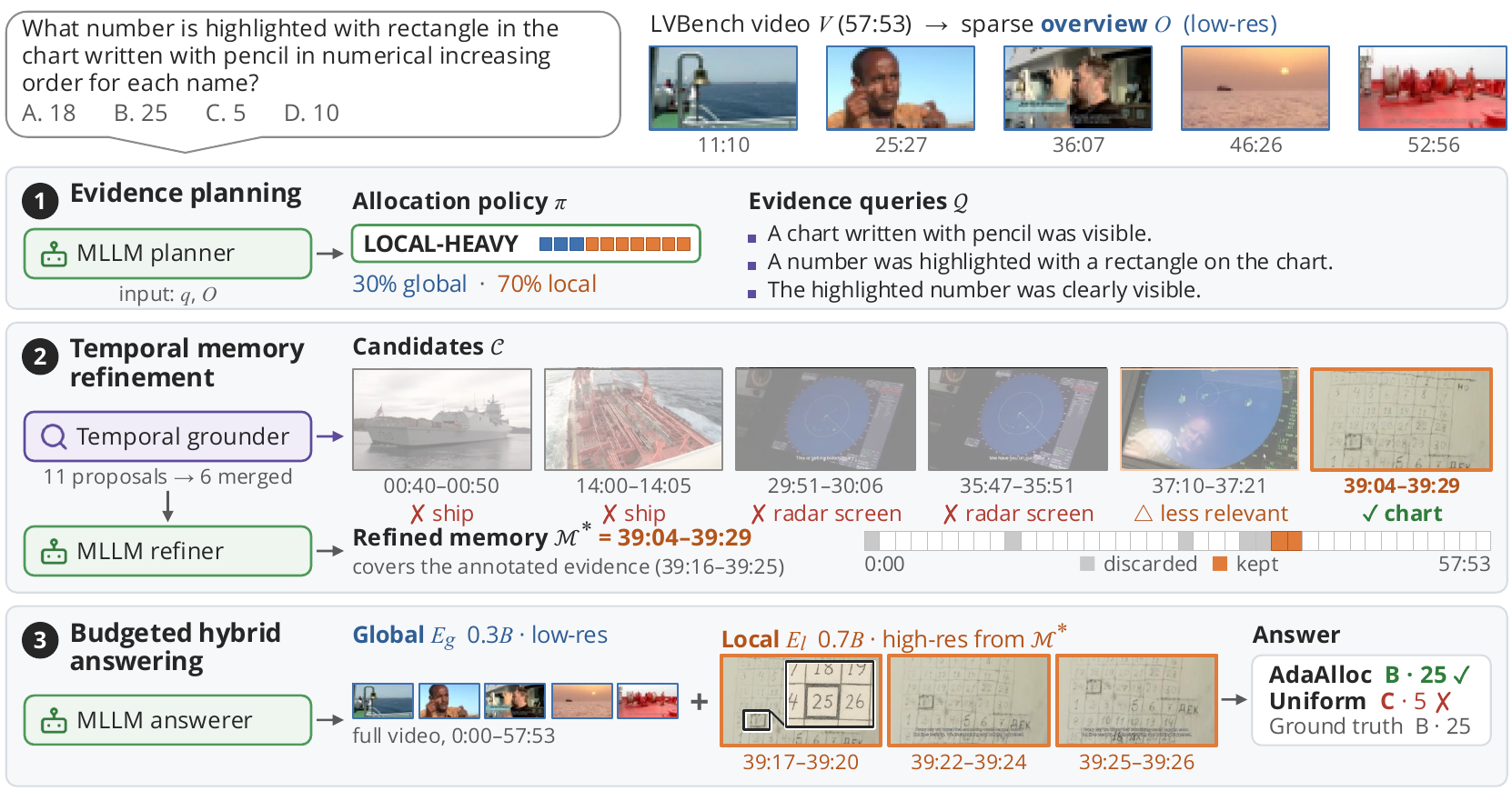}
  \vspace{-4mm}
  \caption{\textbf{Qualitative example on LVBench.}
    The overview supplies global context, while the question's request to read a handwritten number motivates a \textsc{Local-heavy} policy for detailed inspection. AdaAlloc retrieves multiple temporal candidates and refines them into a compact memory that preserves the decisive interval. AdaAlloc correctly answers \textbf{25}, while Uniform predicts~\textbf{5}.
  }
  \label{fig:qualitative}
  \vspace{-3mm}
\end{figure}

\paragraph{Qualitative analysis.}
Figure~\ref{fig:qualitative} illustrates how AdaAlloc allocates evidence on a fine-grained LVBench example. The question asks for the number highlighted by a rectangle in a pencil-written chart, requiring localized OCR/detail perception within a nearly hour-long video.
Conditioned on the question and sparse overview, the planner selects a \textsc{Local-heavy} policy to prioritize detailed inspection of the chart and generates queries targeting the highlighted number.
VTG retrieves multiple candidate intervals, including unrelated distractors such as ship, map, and radar-screen regions, as well as a partial chart-like segment. Temporal-memory refinement filters these noisy proposals and keeps the decisive interval at 00:39:04--00:39:29, where the hand-drawn chart and highlighted number are visible. The final hybrid input retains low-resolution global context while spending high-resolution local tokens on the refined chart region, enabling AdaAlloc to answer correctly with ``25,'' whereas uniform sampling predicts ``5.''

\paragraph{Limitations and future work.}
This work studies training-free visual-token allocation using five discrete global/local policies. This choice simplifies planning and performs comparably to direct ratio prediction on average (Appendix~\ref{sec:appendix_robustness}), though the selected split need not be optimal for each video--question pair. Exploring how to learn or refine the allocation space is a natural next step.
AdaAlloc also relies on a sparse overview and retrieved temporal candidates, so evidence missed by both stages may remain unavailable to memory refinement. Our overview-density analysis suggests limited aggregate sensitivity to sparse planning input (Table~\ref{tab:overview_density}), while motivating further study of evidence recovery in individual failure cases. The additional planning and refinement calls introduce overhead relative to single-pass inference, making the accuracy--latency tradeoff another direction for improvement (Appendix~\ref{sec:appendix_cost}). Finally, our evaluation focuses on frame- and clip-level allocation with 7--8B backbones. Extending allocation to individual visual tokens and studying larger MLLMs would broaden this investigation.

\section{Conclusion}

We introduced AdaAlloc, a training-free agentic framework for long-video understanding that moves beyond frame selection to question-conditioned visual-token allocation. Through overview-conditioned planning, temporal grounding, and memory refinement, it dynamically divides a fixed budget between low-resolution global coverage and high-resolution local evidence, and determines where to spend the local share. Across four long-video QA benchmarks, AdaAlloc consistently improves over budget-matched uniform sampling at 16K and 32K tokens, demonstrating the value of allocating visual capacity across both time and resolution.

\begin{ack}
This research is supported in part by the UT Austin IFML NSF AI Institute and Lyda Hill Philanthropies. We thank Elias Stengel-Eskin and the UT Austin Vision Group for constructive feedback and helpful discussions.
\end{ack}

\bibliographystyle{plainnat}
\bibliography{main}

\clearpage
\appendix
\raggedbottom

\section{MLVU-Dev Results}
\label{sec:appendix_mlvu_dev}

\paragraph{MLVU-Dev results.}
Table~\ref{tab:mlvu_dev} reports results on MLVU-Dev~\citep{zhou2025mlvu}, the development split of MLVU that several prior works report in place of MLVU-Test. It contains 2,174 four-way multiple-choice questions over 1,242 videos in seven tasks. Following the official evaluation, we report M-Avg, the mean of the per-task accuracies; the MLVU-Test results in Table~\ref{tab:main_results} are accuracy over all 502 questions. We use the same Qwen3-VL-8B-Instruct backbone, budgets, and AdaAlloc settings as in Table~\ref{tab:main_results}. AdaAlloc improves M-Avg over uniform sampling by 5.2 points at 16K and 4.0 points at 32K, and improves all seven tasks at both budgets. The largest gains are on Action Count (+12.6 at 16K, +9.3 at 32K) and, at 32K, Action Order (+8.1), the multi-detail tasks that require collecting several pieces of evidence spread across the video. Gains on Topic Reasoning are modest (+1.1 and +1.2), as performance is near ceiling.

\begin{table}[H]
\centering
\small
\setlength{\tabcolsep}{4pt}
\renewcommand{\arraystretch}{1.08}
\caption{
\textbf{MLVU-Dev results.}
Accuracy (\%) on the seven multiple-choice tasks of MLVU-Dev, grouped as in MLVU~\citep{zhou2025mlvu}: Topic Reasoning (TR), Anomaly Recognition (AR), Needle QA (NQA), Ego Reasoning (ER), Plot QA (PQA), Action Order (AO), and Action Count (AC). M-Avg is the mean over the seven tasks.
All methods use Qwen3-VL-8B-Instruct. Numbers in parentheses denote absolute gains over uniform sampling under the same budget.
}
\label{tab:mlvu_dev}
\begin{tabular*}{\textwidth}{@{\extracolsep{\fill}}llcccccccc@{}}
\toprule
\multirow{2}{*}{\textbf{Budget}} & \multirow{2}{*}{\textbf{Method}}
& \multicolumn{2}{c}{\textbf{Holistic}}
& \multicolumn{3}{c}{\textbf{Single-detail}}
& \multicolumn{2}{c}{\textbf{Multi-detail}}
& \multirow{2}{*}{\textbf{M-Avg}} \\
\cmidrule(lr){3-4} \cmidrule(lr){5-7} \cmidrule(lr){8-9}
& & \textbf{TR} & \textbf{AR} & \textbf{NQA} & \textbf{ER} & \textbf{PQA} & \textbf{AO} & \textbf{AC} & \\
\midrule
\multirow{2}{*}{16K tokens}
& Uniform sampling
& 91.3 & 76.0 & 76.9 & 63.4 & 75.0 & 62.9 & 40.3 & 69.4 \\
& \textbf{AdaAlloc}
& \textbf{92.4} & \textbf{78.0} & \textbf{83.1} & \textbf{67.0} & \textbf{80.3} & \textbf{68.3} & \textbf{52.9}
& \textbf{74.6} \textcolor{teal}{\scriptsize (+5.2)} \\
\midrule
\multirow{2}{*}{32K tokens}
& Uniform sampling
& 92.0 & 76.5 & 81.4 & 66.5 & 78.3 & 68.3 & 49.0 & 73.2 \\
& \textbf{AdaAlloc}
& \textbf{93.2} & \textbf{79.5} & \textbf{83.9} & \textbf{67.3} & \textbf{81.4} & \textbf{76.4} & \textbf{58.3}
& \textbf{77.2} \textcolor{teal}{\scriptsize (+4.0)} \\
\bottomrule
\end{tabular*}
\end{table}

\FloatBarrier
\section{VideoMME Breakdown}
\label{sec:appendix_videomme}

\paragraph{VideoMME duration-split results.}
Table~\ref{tab:videomme_duration_breakdown} reports AdaAlloc results on VideoMME without subtitles across the official short, medium, and long duration splits. The short, medium, and long buckets correspond roughly to videos of 0--2 minutes, 4--18 minutes, and 30--60 minutes, respectively. Performance decreases as video length increases, reflecting the growing difficulty of preserving sufficient temporal coverage under a fixed visual-token budget. Increasing the budget from 16K to 32K leaves short videos unchanged at 79.4 and improves medium videos from 68.7 to 70.6 and long videos from 57.2 to 62.2. The gain grows with video length, suggesting that additional visual capacity is especially useful for longer videos.

\begin{table}[H]
\centering
\small
\setlength{\tabcolsep}{8pt}
\renewcommand{\arraystretch}{1.12}
\caption{
\textbf{VideoMME duration-split breakdown.}
We report AdaAlloc performance on VideoMME without subtitles under 16K and 32K visual-token budgets.
VideoMME groups videos into short, medium, and long duration buckets.
}
\label{tab:videomme_duration_breakdown}
\begin{tabular}{lcccc}
\toprule
\multirow{2}{*}{\textbf{Budget}}
& \multicolumn{4}{c}{\textbf{VideoMME$_{\text{w/o sub}}$}} \\
\cmidrule(lr){2-5}
& \textbf{Short} & \textbf{Medium} & \textbf{Long} & \textbf{Overall} \\
\midrule
16K & 79.4 & 68.7 & 57.2 & 68.4 \\
32K & 79.4 & 70.6 & 62.2 & 70.7 \\
\bottomrule
\end{tabular}
\end{table}

\paragraph{VideoMME sub-task breakdown.}
Table~\ref{tab:videomme_subtasks} compares AdaAlloc with uniform sampling (32K) across the official VideoMME sub-tasks. AdaAlloc improves 7 of 12 sub-tasks, with the largest gains on Temporal Reasoning (+5.7) and Spatial Perception (+5.6). Four of the five regressions are within one point. The fifth, Temporal Perception, is the sub-task where uniform sampling already scores highest (85.5), leaving little headroom, and contains only 55 questions, so its $-$5.5 change corresponds to three questions. Counting remains the most challenging category for both methods, likely because it often requires exhaustive visual coverage and reliable aggregation across many frames or repeated instances.

\begin{table}[H]
\centering
\small
\setlength{\tabcolsep}{6pt}
\renewcommand{\arraystretch}{1.08}
\caption{
\textbf{VideoMME sub-task breakdown.}
Accuracy (\%) of uniform sampling and AdaAlloc on VideoMME without subtitles under the same 32K visual-token budget, broken down by the official sub-task categories. \#Q is the number of questions in each sub-task.
}
\label{tab:videomme_subtasks}
\begin{tabular}{lcccc}
\toprule
\textbf{Sub-task} & \textbf{\#Q} & \textbf{Uniform} & \textbf{AdaAlloc} & \textbf{$\Delta$} \\
\midrule
Information Synopsis & 323 & 79.6 & 82.4 & +2.8 \\
Spatial Reasoning & 56 & 78.6 & 80.4 & +1.8 \\
Attribute Perception & 222 & 82.0 & 81.1 & $-$0.9 \\
Object Recognition & 354 & 73.4 & 76.8 & +3.4 \\
Temporal Perception & 55 & 85.5 & 80.0 & $-$5.5 \\
OCR Problems & 139 & 77.7 & 79.9 & +2.2 \\
Spatial Perception & 54 & 68.5 & 74.1 & +5.6 \\
Action Recognition & 313 & 70.9 & 70.6 & $-$0.3 \\
Object Reasoning & 454 & 68.1 & 67.8 & $-$0.3 \\
Temporal Reasoning & 177 & 55.9 & 61.6 & +5.7 \\
Action Reasoning & 285 & 63.5 & 62.8 & $-$0.7 \\
Counting Problem & 268 & 48.9 & 50.4 & +1.5 \\
\midrule
Overall & 2,700 & 69.5 & 70.7 & +1.2 \\
\bottomrule
\end{tabular}
\end{table}

\FloatBarrier
\section{Component Ablation at 16K}
\label{sec:appendix_ablation16k}

Table~\ref{tab:ablation_16k} complements the 32K ablation in Table~\ref{tab:ablation} with a 16K ablation that separates three components: overview-conditioned planning with memory refinement, the global/local hybrid input, and adaptive allocation. Retrieval-only grounds the original question with UniversalVTG and answers from frames sampled from the retrieved segments, without planning or refinement. Local-only runs the full AdaAlloc pipeline but assigns the entire budget to local evidence, and Fixed policy always uses the \textsc{Balanced} split.

\begin{table}[H]
\centering
\small
\setlength{\tabcolsep}{4pt}
\renewcommand{\arraystretch}{1.08}
\caption{
\textbf{Component ablation under the 16K visual-token budget.}
All variants use Qwen3-VL-8B-Instruct with the same final-answering visual-token budget and answerer. Planning \& refinement denotes overview-conditioned planning with temporal-memory refinement; variants that use it share the same overview planner, grounding model, and refinement procedure.
}
\label{tab:ablation_16k}
\resizebox{\textwidth}{!}{
\begin{tabular}{lccccccc}
\toprule
\textbf{Method}
& \makecell{\textbf{Planning \&}\\\textbf{Refinement}}
& \makecell{\textbf{Visual}\\\textbf{Input}}
& \makecell{\textbf{Adaptive}\\\textbf{Allocation}}
& \textbf{LVBench}
& \textbf{MLVU-Test}
& \textbf{VideoMME$_{\text{w/o sub}}$}
& \textbf{LongVideoBench-Val} \\
\midrule
Uniform sampling & $\times$ & Global only & $\times$ & 42.7 & 55.4 & 67.1 & 62.8 \\
Retrieval-only & $\times$ & Local only & $\times$ & 46.5 & 58.8 & 65.9 & 62.5 \\
Local-only & \checkmark & Local only & $\times$ & 48.9 & 58.0 & 66.1 & 64.8 \\
Fixed policy & \checkmark & Global + local & $\times$ & 49.6 & 58.0 & 66.9 & 63.4 \\
\midrule
\textbf{AdaAlloc} & \checkmark & Global + local & \checkmark & \textbf{51.7} & \textbf{61.2} & \textbf{68.4} & \textbf{66.0} \\
\bottomrule
\end{tabular}
}
\end{table}

Adding overview-conditioned planning and memory refinement on top of question-only retrieval (Retrieval-only vs.\ Local-only) improves LVBench and LongVideoBench by 2.4 and 2.3 points and VideoMME by 0.2 points, but lowers MLVU-Test by 0.8 points. Adding a fixed global stream (Local-only vs.\ Fixed policy) then changes accuracy by only +0.7, +0.0, +0.8, and $-$1.4 points, so the global/local hybrid representation alone does not explain AdaAlloc's gains. Finally, Fixed policy and AdaAlloc share every component and differ only in whether the global/local split is fixed or selected for each video--question pair. Adaptive allocation improves all four benchmarks, by +2.1, +3.2, +1.5, and +2.6 points, and the same comparison at 32K (Table~\ref{tab:ablation}) shows gains of 1.6--6.5 points, so question-conditioned allocation provides a consistent improvement beyond combining global and local evidence.

\FloatBarrier
\section{Robustness and Design Analysis}
\label{sec:appendix_robustness}

\paragraph{Sensitivity to overview density.}
The planner observes only a sparse overview, so subtle local evidence may be absent from its input, and planning errors could propagate to later stages. To measure the end-to-end effect of this information loss, we keep the full pipeline and the 16K final answering budget unchanged and vary only the planner's visual input: the default overview versus a $2\times$ denser overview with twice as many uniformly sampled frames at the same resolution. As shown in Table~\ref{tab:overview_density}, the two conditions differ only marginally, with two benchmarks improving and two declining slightly. Doubling the planner's visual input thus yields no consistent benefit, suggesting that the default overview is sufficient for the planner's decisions in aggregate. This analysis does not eliminate the information loss inherent to any finite overview: individual planning errors can still occur, and when the overview and grounding model both miss the decisive event, later stages may fail to recover it.
The 16K component ablation (Table~\ref{tab:ablation_16k}) provides complementary evidence: adding overview-conditioned planning and memory refinement on top of question-only retrieval (Retrieval-only vs.\ Local-only), with the retriever, local budget, and final answerer fixed, improves three of the four benchmarks and the average (+1.0 points), so the overview-conditioned stages improve rather than degrade end-to-end accuracy despite the sparse overview.

\begin{table}[H]
\centering
\small
\setlength{\tabcolsep}{4pt}
\renewcommand{\arraystretch}{1.08}
\caption{
\textbf{Sensitivity to overview density.}
The final answering budget is fixed at 16K and only the planner's visual input changes. The $2\times$ denser overview samples twice as many frames at the same resolution and spends additional planning compute, so it is a diagnostic rather than a budget-matched alternative.
}
\label{tab:overview_density}
\begin{tabular*}{\textwidth}{@{\extracolsep{\fill}}lcccc@{}}
\toprule
\textbf{Planner Overview}
& \textbf{LVBench}
& \textbf{MLVU-Test}
& \textbf{VideoMME$_{\text{w/o sub}}$}
& \textbf{LongVideoBench-Val} \\
\midrule
Default & 51.7 & 61.2 & 68.4 & 66.0 \\
$2\times$ denser & 51.6 & 61.4 & 68.9 & 65.8 \\
\midrule
Difference & $-$0.1 & +0.2 & +0.5 & $-$0.2 \\
\bottomrule
\end{tabular*}
\end{table}

\paragraph{Discrete policies versus direct ratio prediction.}
AdaAlloc's planner selects among five predefined policies. To test whether this discrete design limits performance, we compare it with a variant in which the planner directly predicts the global/local ratio. As shown in Table~\ref{tab:direct_ratio}, the two variants are nearly tied at 16K, while at 32K policy selection is better on LVBench and LongVideoBench, ties on MLVU-Test and VideoMME, and leads by 0.57 points on average. Allowing the planner to output an arbitrary ratio therefore does not consistently improve performance. The discrete policies provide well-separated operating points across the global-to-local spectrum and reduce the planner's decision from predicting a precise continuous value to choosing among a few clear alternatives, while remaining adaptive to each video--question pair. We do not claim that these ratios are optimal; rather, the results indicate that the discrete policy set is a reasonable and stable design.

\begin{table}[H]
\centering
\small
\setlength{\tabcolsep}{3pt}
\renewcommand{\arraystretch}{1.08}
\caption{
\textbf{Discrete policy selection versus direct ratio prediction.}
Both variants use the same planner, grounding model, token budget, and answering pipeline, and differ only in whether the planner selects one of the five predefined policies or directly predicts the global/local ratio.
}
\label{tab:direct_ratio}
\begin{tabular*}{\textwidth}{@{\extracolsep{\fill}}llccccc@{}}
\toprule
\textbf{Budget} & \textbf{Allocation}
& \textbf{LVBench}
& \textbf{MLVU-Test}
& \textbf{VideoMME$_{\text{w/o sub}}$}
& \textbf{LongVideoBench-Val}
& \textbf{Avg.} \\
\midrule
\multirow{2}{*}{16K}
& Direct ratio prediction & 51.1 & \textbf{61.4} & \textbf{68.8} & 65.5 & 61.68 \\
& Policy selection (ours) & \textbf{51.7} & 61.2 & 68.4 & \textbf{66.0} & \textbf{61.82} \\
\midrule
\multirow{2}{*}{32K}
& Direct ratio prediction & 54.9 & \textbf{67.1} & \textbf{70.7} & 67.0 & 64.94 \\
& Policy selection (ours) & \textbf{56.6} & \textbf{67.1} & \textbf{70.7} & \textbf{67.5} & \textbf{65.51} \\
\bottomrule
\end{tabular*}
\end{table}

\paragraph{Sensitivity to the temporal retrieval module.}
AdaAlloc only requires a module that maps planner-generated queries to relevant video moments. We use UniversalVTG~\citep{an2026universalvtg} by default because it is an efficient open-vocabulary grounder that directly returns ranked temporal intervals. As an alternative, we use query--frame similarity retrieval, a common mechanism in long-video QA and frame-selection methods~\citep{chen2026eventanchored,liu2025bolt,luo2025videorag,sun2025mdp3,tang2025adaptive,wang2026wheretofocus,zhang2025qframe,zhu2025focus}, which ranks frames by the cosine similarity between each planner-generated query and the frames, both embedded with the Perception Encoder (PE-Core) image--text model~\citep{bolya2025perception}.
As shown in Table~\ref{tab:retriever_ablation}, AdaAlloc with query--frame similarity still outperforms uniform sampling on every benchmark; its average is 0.31 points above UniversalVTG at 16K and 1.66 points below it at 32K. We hypothesize that this reflects how the two retrievers use a small local budget: query--frame similarity selects individual high-scoring frames, which is efficient when few local frames are available, whereas UniversalVTG returns contiguous intervals whose temporal context may be better captured by the denser sampling available at 32K. UniversalVTG is therefore our default, but AdaAlloc retains most of its gains with a simpler and commonly used retrieval mechanism, indicating that its improvements do not depend on a specific grounding model.

\begin{table}[H]
\centering
\small
\setlength{\tabcolsep}{3pt}
\renewcommand{\arraystretch}{1.08}
\caption{
\textbf{Sensitivity to the temporal retrieval module.}
AdaAlloc with UniversalVTG~\citep{an2026universalvtg} (default) or query--frame similarity retrieval, with the planner, allocation policy, token budget, memory refinement, and final answerer unchanged.
}
\label{tab:retriever_ablation}
\begin{tabular*}{\textwidth}{@{\extracolsep{\fill}}llccccc@{}}
\toprule
\textbf{Budget} & \textbf{Retriever}
& \textbf{LVBench}
& \textbf{MLVU-Test}
& \textbf{VideoMME$_{\text{w/o sub}}$}
& \textbf{LongVideoBench-Val}
& \textbf{Avg.} \\
\midrule
\multirow{3}{*}{16K}
& Uniform sampling & 42.7 & 55.4 & 67.1 & 62.8 & 56.99 \\
& Query--frame similarity & \textbf{53.3} & \textbf{61.8} & \textbf{69.9} & 63.5 & \textbf{62.13} \\
& UniversalVTG & 51.7 & 61.2 & 68.4 & \textbf{66.0} & 61.82 \\
\midrule
\multirow{3}{*}{32K}
& Uniform sampling & 46.1 & 61.4 & 69.5 & 63.9 & 60.21 \\
& Query--frame similarity & 51.9 & \textbf{67.3} & 70.5 & 65.7 & 63.85 \\
& UniversalVTG & \textbf{56.6} & 67.1 & \textbf{70.7} & \textbf{67.5} & \textbf{65.51} \\
\bottomrule
\end{tabular*}
\end{table}

\paragraph{Generalization to another MLLM backbone.}
To test whether the gains are specific to Qwen3-VL, we apply AdaAlloc to VideoLLaMA3-7B~\citep{zhang2025videollama3}, which differs from Qwen3-VL in both vision encoder and training recipe. We use the same evaluation protocol and a matched 32K visual-token budget for AdaAlloc and uniform sampling. As shown in Table~\ref{tab:backbone_generalization}, AdaAlloc improves VideoLLaMA3-7B on all four benchmarks, with an average gain of 2.25 points, indicating that its benefits are not tied to a single backbone or model family.

\begin{table}[H]
\centering
\small
\setlength{\tabcolsep}{4pt}
\renewcommand{\arraystretch}{1.08}
\caption{
\textbf{Generalization to another MLLM backbone.}
Accuracy (\%) with VideoLLaMA3-7B~\citep{zhang2025videollama3} under a matched 32K visual-token budget.
Numbers in parentheses denote absolute gains over uniform sampling.
}
\label{tab:backbone_generalization}
\begin{tabular*}{\textwidth}{@{\extracolsep{\fill}}lccccc@{}}
\toprule
\textbf{Method}
& \textbf{LVBench}
& \textbf{MLVU-Test}
& \textbf{VideoMME$_{\text{w/o sub}}$}
& \textbf{LongVideoBench-Val}
& \textbf{Avg.} \\
\midrule
Uniform sampling
& 43.4 & 56.2 & 64.8 & 61.0 & 56.35 \\
\textbf{AdaAlloc}
& \textbf{47.2} \textcolor{teal}{\scriptsize (+3.8)}
& \textbf{58.4} \textcolor{teal}{\scriptsize (+2.2)}
& \textbf{65.9} \textcolor{teal}{\scriptsize (+1.1)}
& \textbf{62.9} \textcolor{teal}{\scriptsize (+1.9)}
& \textbf{58.60} \textcolor{teal}{\scriptsize (+2.25)} \\
\bottomrule
\end{tabular*}
\end{table}

\FloatBarrier
\section{Allocation Policy Analysis}
\label{sec:appendix_policy}

\begin{figure}[H]
    \centering
    \includegraphics[width=\linewidth]{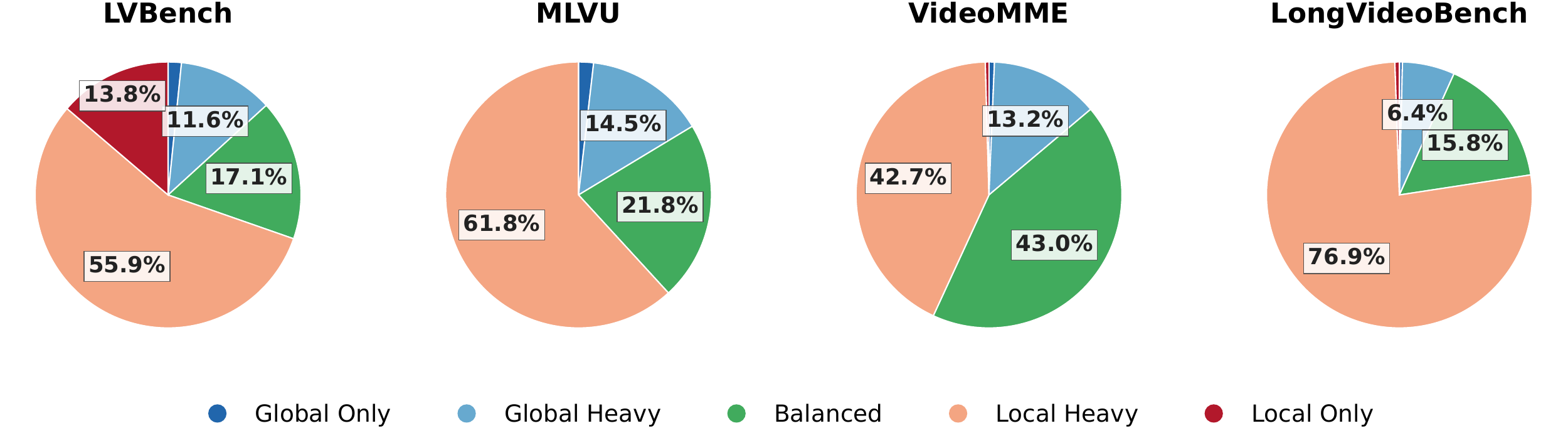}
    \caption{\textbf{Distribution of selected allocation policies across benchmarks.}
    We report the fraction of planner-routed examples assigned to each AdaAlloc policy on LVBench, MLVU, VideoMME, and LongVideoBench.
    Examples that bypass planning and use direct answering are excluded.
    }
    \label{fig:policy_distribution}
\end{figure}

\paragraph{Distribution of selected allocation policies.}
Figure~\ref{fig:policy_distribution} shows the distribution of allocation policies selected by AdaAlloc across benchmarks. AdaAlloc first checks whether uniform full-video sampling under the assigned visual-token budget already provides sufficiently dense temporal coverage. If so, it bypasses temporal retrieval and memory refinement, and directly answers from the global overview. Otherwise, the planner selects among five policies: global-only, global-heavy, balanced, local-heavy, and local-only. Across most benchmarks, AdaAlloc favors policies that retain global context while allocating substantial budget to localized evidence. In particular, local-heavy allocation is selected most frequently on LVBench, MLVU, and LongVideoBench, suggesting that many long-video questions require targeted high-resolution inspection after a sparse overview. VideoMME shows a more balanced distribution between balanced and local-heavy policies, indicating a larger fraction of examples where broad temporal coverage and localized verification are both important. Local-only allocation is uncommon outside LVBench, while mixed global/local policies dominate across all four benchmarks. This suggests that preserving global context is often useful even when local evidence is needed.

\paragraph{Policy selection by question category.}
Existing QA benchmarks do not provide a ground-truth global/local ratio, so policy accuracy is not directly defined. We instead examine whether the selected policies follow interpretable patterns, using the question-family metadata of NExT-QA~\citep{xiao2021next}, which groups questions into descriptive, causal, and temporal families. Table~\ref{tab:policy_by_category_nextqa} reports the selected policies on the validation split under the 16K budget. Descriptive questions select global-oriented policies (\textsc{Global-only} or \textsc{Global-heavy}) more often than the other families, causal questions shift toward \textsc{Balanced}, and temporal questions have the highest \textsc{Local-heavy} rate.

\begin{table}[H]
\centering
\small
\setlength{\tabcolsep}{6pt}
\renewcommand{\arraystretch}{1.08}
\caption{
\textbf{Selected policies by NExT-QA question family (\%).}
NExT-QA validation split (4,996 questions) under the 16K budget. No question selected \textsc{Local-only}, so the column is omitted. Question families are benchmark metadata and are not available to the planner. Rows may not sum to 100 due to rounding.
}
\label{tab:policy_by_category_nextqa}
\begin{tabular}{lcccc}
\toprule
\textbf{Question Family}
& \textbf{Global-only}
& \textbf{Global-heavy}
& \textbf{Balanced}
& \textbf{Local-heavy} \\
\midrule
Descriptive & 4.4 & 13.5 & 33.1 & 49.0 \\
Causal & 0.2 & 7.5 & 43.8 & 48.6 \\
Temporal & 0.1 & 8.7 & 38.6 & 52.5 \\
\bottomrule
\end{tabular}
\end{table}

These labels are unavailable at inference. The planner conditions only on the question, answer choices, and sparse overview, so this is a post-hoc interpretability check. Policy choices vary within each family, reflecting adaptation to individual video--question pairs.

\FloatBarrier
\section{Inference Cost}
\label{sec:appendix_cost}

\paragraph{Measurement setup.}
Our accuracy experiments were run on NVIDIA A40 and GH200 GPUs; all cost measurements in this section use NVIDIA A40 GPUs so that methods are compared on the same hardware.
We measure total compute and end-to-end wall-clock runtime for uniform sampling, DIG~\citep{li2025divide}, and AdaAlloc using the same Qwen3-VL-8B-Instruct backbone and matched 16K/32K visual-token budgets.
Each MLLM call, including AdaAlloc's planning and memory-refinement calls, is capped at the same visual-token budget $B$; the reported compute and runtime include all calls.
Uniform sampling, AdaAlloc, and all DIG components except its initial text-only query classification run on a single NVIDIA A40 GPU. DIG's query classification uses Qwen3-Next-80B-A3B-Instruct, which cannot be served on a single A40 and therefore runs on eight A40 GPUs.

\begin{table}[H]
\centering
\small
\setlength{\tabcolsep}{8pt}
\renewcommand{\arraystretch}{1.08}
\caption{
\textbf{Inference cost under matched visual-token budgets.}
Total compute and end-to-end wall-clock runtime of uniform sampling, DIG~\citep{li2025divide}, and AdaAlloc, all with the Qwen3-VL-8B-Instruct backbone.
}
\label{tab:inference_cost}
\begin{tabular}{llcc}
\toprule
\textbf{Budget} & \textbf{Method} & \textbf{Compute (TFLOPs)} & \textbf{Runtime (s)} \\
\midrule
\multirow{3}{*}{16K}
& Uniform sampling & 468 & 14 \\
& DIG~\citep{li2025divide} & 1,871 & 109 \\
& \textbf{AdaAlloc} & 1,146 & 35 \\
\midrule
\multirow{3}{*}{32K}
& Uniform sampling & 1,261 & 25 \\
& DIG~\citep{li2025divide} & 2,664 & 119 \\
& \textbf{AdaAlloc} & 2,873 & 58 \\
\bottomrule
\end{tabular}
\end{table}

\paragraph{Results.}
As shown in Table~\ref{tab:inference_cost}, uniform sampling is the cheapest because it requires a single MLLM call. Compared with DIG, AdaAlloc is $3.1\times$ faster with 39\% fewer FLOPs at 16K, and $2.1\times$ faster with 7.8\% more FLOPs at 32K. DIG's runtime changes little across budgets because its reward-scoring stage introduces a largely budget-independent cost. Overall, AdaAlloc costs more than single-pass uniform sampling but has substantially lower latency than a comparable training-free selection method.

\FloatBarrier
\section{Differences from the Submitted Version}
\label{sec:appendix_changes}

While preparing the code release, we found that our UniversalVTG integration did not correctly apply non-maximum suppression and top-$K$ selection to the retrieved temporal proposals. We fixed this issue and now retain the top-5 proposals per query. We also changed all MLLM calls (planning, memory refinement, and answering) from Qwen3-VL's default sampling-based decoding to greedy decoding, which makes predictions deterministic and is better suited to multiple-choice evaluation. We reran our experiments with these changes, and the reported results of those experiments reflect the updated settings. The method, its components, and our conclusions are unchanged. The retriever fix affects every setting that uses UniversalVTG for temporal retrieval, including AdaAlloc and its retrieval-based variants. These changes improve LVBench accuracy (e.g., AdaAlloc at 32K increases from 53.5 to 56.6), with mixed changes on other benchmarks. Some numbers in this version therefore differ from those in the submission and rebuttal.

\FloatBarrier
\section{Broader Impacts}
\label{sec:appendix_broader_impacts}

Improving long-video question answering with existing MLLMs may make video archives easier to search and support accessibility applications. However, missed or misinterpreted evidence can produce incorrect answers, and analyzing personal recordings can create privacy risks or enable intrusive surveillance. Applications should use videos with appropriate authorization and include human review when incorrect answers could have consequential effects.

\end{document}